\documentclass[sigconf]{acmart}
\usepackage{upgreek}
\usepackage{multirow}
\AtBeginDocument{%
  }

\setcopyright{acmlicensed}
\copyrightyear{2025}
\acmYear{2025}
\acmDOI{XXXXXXX.XXXXXXX}
\acmConference[Conference acronym 'XX]{Make sure to enter the correct
  conference title from your rights confirmation email}{2025}{Woodstock, NY}
\acmISBN{978-1-4503-XXXX-X/2018/06}

\begin{document}

%\title{Can-SAVE: Deploying Survival-Based AI for Low-Cost, Large-Scale Cancer Screening}
%\title{Can-SAVE: Mass AI Cancer Screening \\ via Survival Analysis Variables and EHR}
\title{Can-SAVE: Deploying Low-Cost and Population-Scale \\ Cancer Screening via Survival Analysis Variables and EHR}

\author{Petr Philonenko}
\affiliation{%
  \institution{Sber AI Lab}
  \city{Moscow}
  \country{Russia}}
\email{petr-filonenko@mail.ru}

\author{Vladimir Kokh}
\affiliation{%
  \institution{Sber AI}
  \city{Moscow}
  \country{Russia}}
\email{kokh.v.n@sber.ru}

\author{Pavel Blinov}
\affiliation{%
  \institution{Sber AI Lab}
  \city{Moscow}
  \country{Russia}}
\email{blinov.p.d@sber.ru}

\renewcommand{\shortauthors}{Philonenko et al.}

\begin{abstract}
Conventional medical cancer screening methods are costly, labor-intensive, and extremely difficult to scale. Although AI can improve cancer detection, most systems rely on complex or specialized medical data, making them impractical for large-scale screening. We introduce Can-SAVE, a lightweight AI system that ranks population-wide cancer risks solely based on medical history events. By integrating survival model outputs into a gradient-boosting framework, our approach detects subtle, long-term patient risk patterns -- often well before clinical symptoms manifest. Can-SAVE was rigorously evaluated on a real-world dataset of 2.5 million adults spanning five Russian regions, marking the study as one of the largest and most comprehensive deployments of AI-driven cancer risk assessment. In a retrospective oncologist-supervised study over 1.9M patients, Can-SAVE achieves a 4–10x higher detection rate at identical screening volumes and an Average Precision (AP) of 0.228 vs. 0.193 for the best baseline (LoRA-tuned Qwen3-Embeddings via DeepSeek-R1 summarization). In a year-long prospective pilot (426K patients), our method almost doubled the cancer detection rate (+91\%) and increased population coverage by 36\% over the national screening protocol. The system demonstrates practical scalability: a city-wide population of 1 million patients can be processed in under three hours using standard hardware, enabling seamless clinical integration. This work proves that Can-SAVE achieves nationally significant cancer detection improvements while adhering to real-world public healthcare constraints, offering immediate clinical utility and a replicable framework for population-wide screening. Code for training and feature engineering is available at \texttt{https://github.com/sb-ai-lab/Can-SAVE}.

\end{abstract}




%   \includegraphics[width=\textwidth]{TEASER.png}
%   \caption{Intro.}

\received{20 February 2007}
\received[revised]{12 March 2009}
\received[accepted]{5 June 2009}

\maketitle

\section{Introduction}
Cancer screening represents one of healthcare's most persistent challenges~\cite{screenings_challenges}, with traditional methods often proving too costly and resource-intensive for population-wide implementation~\cite{screenings_cost}. While early cancer detection significantly improves survival times, existing screening programs typically achieve modest detection rates. For example, the best screening for colorectal cancer identifies only up to 9 cases per 1,000 examinations~\cite{NNS_colorectal}. This creates a critical gap between the need for screening and the capacity of the healthcare system, particularly in resource-limited settings.

The proliferation of Electronic Health Records (EHR)~\cite{EHR} offers unprecedented opportunities to address this challenge through automated risk stratification. However, most existing AI approaches for cancer prediction require specialized data (e.g. genetic data~\cite{genetic_data}, biomarkers~\cite{biomarkers}, family history~\cite{family_history}, lifestyle~\cite{lifestyle}, bad habits~\cite{bad_habits}, interactions with harmful substances~\cite{interactions_with_harmful_substances}, etc.), extensive preprocessing, or complex infrastructure that limits their practical deployment. Previous research has focused primarily on algorithmic performance rather than real-world implementation challenges, creating a disconnect between theoretical advances and clinical utility~\cite{ai_models_harder_rather_applicable}.

Healthcare systems worldwide face similar constraints: limited budgets, data quality, and the need for immediately deployable solutions. These constraints require a different approach to AI development. Our work addresses this gap by demonstrating how survival models can be combined with standard machine learning techniques to create effective and deployable cancer risk prediction systems.

Notably, although significant progress has been made in cancer risk prediction, to our knowledge, no existing solution fully addresses the challenge of scalable, low-resource, population-scale cancer screening using only routine EHR and medical service codes. This positions our work as the first to demonstrate a practical, infrastructure-agnostic system grounded in survival models and standard ML techniques for nationwide, real-world deployment.

This paper presents comprehensive insights into the implementation of cancer risk prediction in 2.5 million patients in five Russian regions, representing one of the largest real-world validations of EHR-based cancer prediction. We emphasize practical lessons learned from large-scale deployment, including challenges in system integration, adaptation of clinical workflow, and performance validation in different populations.

%Our work makes several contributions to applied healthcare:
Our work bridges data science and healthcare through the following key contributions:
\begin{enumerate}
    \item Demonstrate how survival models can enhance traditional machine learning approaches using only routine EHR data, requiring no specialized infrastructure or data collection; 
    \item Provide comprehensive validation across 2.5 million patients, including real-world 12-month prospective experiment supervised by oncologists; 
    \item Present a scalable solution that can be adapted in different healthcare systems and EHR formats;
    \item Quantify the post-deployment performance of the system, demonstrating a 91\% increase in cancer detection rate and a 36 percentage-point expansion in population coverage relative to the national protocol.
\end{enumerate}

By enabling early and low-cost risk stratification, Can-SAVE empowers healthcare systems to optimize screening resources, reduce late-stage diagnoses, and ultimately save lives. Its minimal data requirements make it accessible to clinics worldwide, even in resource-limited settings, ushering in a new paradigm for population-scale cancer prevention.

%The remainder of this paper presents our methodology, extensive validation results, and deployment description that can guide similar implementations in other healthcare settings.

% ----------------

\section{Related Work}

\noindent\textbf{General Problem Formulation}

The challenge of population-level cancer screening represents one of the most pressing implementation problems in healthcare. Traditional methods often prove costly, time-consuming, and poorly suited for large-scale deployment. While specialized medical parameters can increase the sensitivity of the model to the target disease (cancer), they significantly narrow their applicability to mass screening~\cite{complex_data_for_cancer}. The proliferation of EHR offers unprecedented opportunities for automated risk stratification, yet most existing AI approaches require extensive preprocessing or complex infrastructure that limits practical deployment~\cite{stop_scalling}. 
%A solution to a similar problem was raised in the work~\cite{template_nature}, however, this study is also limited to only predicting pancreatic cancer.
While prior work~\cite{template_nature} addresses a similar problem, it focuses exclusively on pancreatic cancer prediction, limiting its applicability to broader population-wide screening.
Similarly, the MEDomics framework addresses pan-cancer prognostication through continuous learning from longitudinal EHR data~\cite{medomics}, though it focuses on survival prediction for diagnosed patients rather than population-wide screening, and requires complex multimodal data infrastructure including imaging and natural language processing capabilities.

These implementation gaps necessitate a fundamentally different solution -- one that prioritizes practical applicability over theoretical complexity. Our work addresses this challenge by demonstrating how established techniques can be strategically combined to create effective and deployable solutions for resource-constrained healthcare settings.

\noindent\textbf{Machine Learning Methods}

Classical machine learning approaches have shown substantial success in cancer risk prediction, particularly when deployed in resource-limited settings. Gradient boosting methods have become particularly effective for healthcare applications due to their superior handling of tabular data and native support for categorical features~\cite{ml_best}. Recent studies demonstrate that CatBoost~\cite{CatBoost} and similar gradient boosting frameworks consistently outperform traditional methods across diverse healthcare prediction tasks~\cite{catboost_best}. The interpretability and computational efficiency of these methods make them ideal candidates for large-scale deployment scenarios.

Random Forest~\cite{breiman2001random} approaches have shown remarkable performance in cancer risk assessment, particularly in breast cancer prediction, where they achieve accuracy rates that exceed 90\% on diverse patient populations~\cite{RF_90perc}. The ensemble nature of Random Forest provides robust handling of missing data and feature interactions, critical considerations for real-world EHR deployment. Studies demonstrate that Random Forest maintains consistent performance in different health systems and patient demographics~\cite{RF_stable}.

Logistic regression remains fundamental for healthcare risk prediction due to its interpretability and regulatory compliance advantages~\cite{LogReg_forever}. The linear nature of logistic regression enables straightforward feature importance analysis and clinical decision support, essential requirements for healthcare deployment. Recent work shows that well-engineered logistic regression models can achieve competitive performance with more complex approaches while maintaining the transparency required for clinical adoption~\cite{LogReg_interpretable}.

\noindent\textbf{Survival Analysis Methods}

Survival analysis models have gained significant traction in healthcare applications, particularly for cancer prognosis and risk stratification. Accelerated Failure Time (AFT) models provide intuitive interpretations of covariate effects, directly modeling the acceleration or deceleration of time-to-event outcomes~\cite{AFT_overview}. Recent research of AFT models demonstrate substantial improvements over traditional approaches, achieving a better fit of the model while maintaining clinical interpretability~\cite{AFT_comparison}. AFT models have proven particularly effective for EHR-based prediction tasks where the time-to-event interpretation provides actionable clinical insights.

Random Survival Forest (RSF) extends the Random Forest framework to handle censored data, providing ensemble-based survival prediction with enhanced robustness~\cite{RSF_application}. RSF models demonstrate superior performance in high-dimensional settings common in healthcare applications, particularly when dealing with complex feature interactions and non-linear relationships~\cite{RSF_comparison}. The variable importance measures provided by RSF enable the identification of key risk factors while maintaining the ensemble robustness that makes Random Forest approaches successful in healthcare settings.

Deep survival analysis methods have emerged as powerful tools for complex survival prediction tasks, particularly for longitudinal EHR data~\cite{DeepHit,DSM}. Recent work demonstrates that neural network-based survival models can achieve superior performance compared to traditional approaches, especially when handling high-dimensional data or complex temporal patterns~\cite{DeepSurvivalAnalysis_comparison}. However, these approaches typically require substantial computational resources and extensive training data, limiting their applicability~\cite{DeepSurvivalAnalysis_heavy} in resource-constrained deployment scenarios.

\noindent\textbf{Deep Learning Methods}

Advanced deep learning architectures have shown promise for healthcare prediction tasks. Fine-tuned clinical language models have demonstrated significant improvements over general-purpose models for medical text analysis and structured data prediction~\cite{DL_FineTuned}. 

Pre-trained BERT~\cite{devlin2019bert} models adapted for medical domains show strong performance in EHR-based prediction tasks, particularly when combined with domain-specific fine-tuning~\cite{BERT_Profile,BERT_Baseline}. The bidirectional nature of BERT enables an effective understanding of the context of medical terminology and clinical relationships~\cite{BERT_Biderection}. Recent work demonstrates that medical BERT models can achieve competitive performance with specialized architectures while maintaining a wider applicability~\cite{BERT_concurent}.

The Longformer architecture~\cite{beltagy2020longformer}, which contains a large context length and was pre-trained on clinical notes, consistently outperforms standard BERT models in multiple healthcare tasks~\cite{Longformer}. However, these models also require substantial computational resources and specialized infrastructure for deployment.

\noindent\textbf{LLM-based Methods}

Large Language Models (LLMs) have recently emerged as powerful tools for healthcare prediction and clinical decision support, though their deployment complexity presents significant challenges for widespread adoption. Clinical prediction with LLMs has shown remarkable performance improvements over traditional methods, with recent work demonstrating that fine-tuned LLMs can significantly outperform state-of-the-art models in both PR-AUC and ROC-AUC metrics~\cite{CPLLM}. Furthermore, recent reviews demonstrate that LLMs can substantially outperform traditional deep survival methods such as DeepHit~\cite{llm_vs_deephit}. The LLM approach eliminates the need for pre-training on clinical data while achieving superior performance in multiple healthcare prediction tasks and providing greater deployment flexibility~\cite{LLM_flexible_deployment}. And a wide context window of modern LLMs enables the processing of extensive medical documents and comprehensive patient histories.

Embedding-based approaches using LLM-derived representations show promise for efficient healthcare prediction while maintaining the semantic understanding advantages of large language models. These methods provide a middle ground between full LLM deployment and traditional feature engineering, offering improved performance with reduced computational requirements. Recent work demonstrates that embedding-based approaches can achieve competitive performance with full LLM fine-tuning.

% ----------------

\section{Methodology}

\begin{figure*}[]
  \includegraphics[width=12cm]{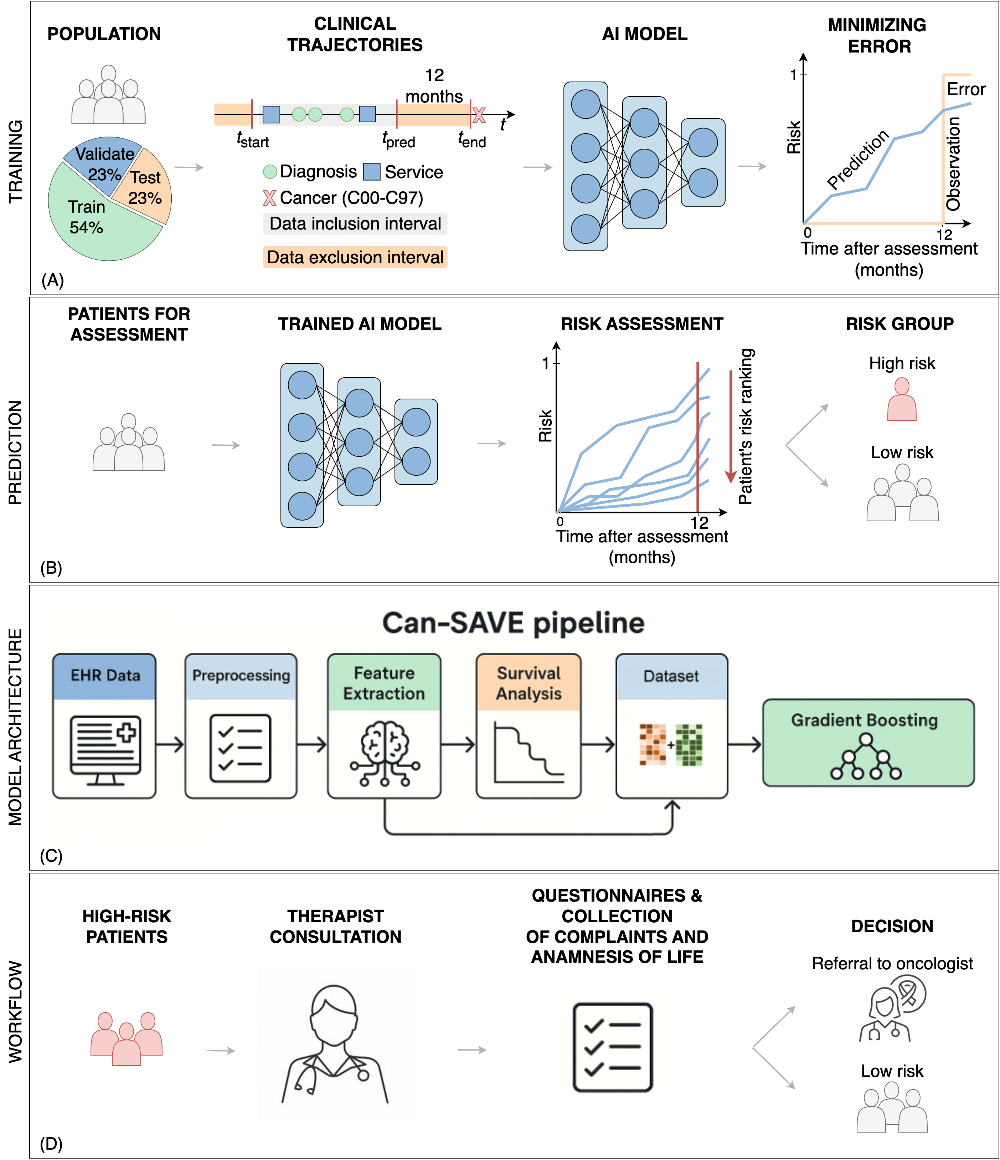}
  \caption{Overview of Can-SAVE}
  \label{fig:overview}
\end{figure*}

\subsection{Problem Formulation}
We formulate the prediction of cancer risk as a \textbf{\textit{binary classification problem}} with temporal considerations. For each patient visit at time $t_{pred}$, we predict the risk of cancer diagnosis within the following 12~months (Figure~\ref{fig:overview}A). Let $Q_i$ represent the EHR of patient $i$, containing chronologically ordered medical events $E_{ij} = (date_{ij}, code_{ij}, type_{ij})$, where $type_{ij}$ indicates either diagnosis or medical service.

The prediction \textbf{\textit{target}} is defined as:
\begin{itemize}
    \item $target = 1$ if cancer diagnosis (ICD-10 C00-C97) occurs within $[t_{pred}, t_{pred} + 12M]$;
    \item $target = 0$ otherwise.
\end{itemize}

This formulation enables direct comparison of risk across patients and supports ranking-based deployment scenarios where healthcare systems must prioritize limited screening resources.

\subsection{Baselines}
To accommodate various types of EHR signal and deployment constraints, we implement six complementary modeling pipelines. Each pipeline follows the same high-level template: extract structured or semantic features from raw events, then estimate
(a) the probability of a cancer diagnosis within the next 12 months \texttt{P(Cancer|EHR)} or 
(b) the time-to-event distribution \texttt{S(t|EHR)}. \\

\noindent\textbf{\textit{Machine Learning Pipeline:}} \\
\texttt{EHR Data → Feature Engineering → Classical ML → P(Cancer|EHR).}  \\
List of ML-based solutions:
\begin{itemize}
    \item Logistic regression;
    \item Random Forest;
    \item Gradient Boosting Machine (GBM).
\end{itemize}

\noindent\textbf{\textit{Survival Models Pipeline:}} \\
\texttt{EHR Data → Survival Features → Survival Models → S(t|EHR).} \\
List of survival model solutions:
\begin{itemize}
    \item Accelerated Failure Time (AFT) model;
    \item Randon Survival Forest;
    \item DeepHit~\cite{DeepHit};
    \item Deep Survival Machines~\cite{DSM}.
\end{itemize}

\noindent\textbf{\textit{Deep Learning (RNN \& Transformers) Pipeline:}} \\
\texttt{EHR Data → Sequence Events Encoding → Temporal Modeling → P(Cancer|EHR).} \\
List of deep learning solutions:
\begin{itemize}
    \item Fine-tuned CoLES~\cite{CoLES} for both Sequence Events Encoding and Temporal Modeling;
    \item Pre-trained BERT (Profile model~\cite{BERT_Profile}) for Sequence Events Encoding and GRU for Temporal Modeling;
    \item Longformer pre-trained on medical texts~\cite{Longformer} for Sequence Events Encoding and CatBoost for binary classification.
\end{itemize}

\noindent\textbf{\textit{LLM Encoding Pipeline:}} \\
\texttt{EHR Data → Medical Text → LLM Encoding → Semantic Features → P(Cancer|EHR) or S(t|EHR).} \\
List of LLM Encoders:
\begin{itemize}
    \item DeepSeek-R1-Distill-Qwen-1.5B (last hidden layer);
    \item Qwen3-Embedding-0.6B;
    \item GigaChat-Embeddings.
\end{itemize}

\noindent\textbf{\textit{LLM Summarization \& Encoding Pipeline:}} \\
\texttt{EHR Data → Medical Text → LLM Summarization → LLM Encoding → Semantic Features → P(Cancer|EHR) or S(t|EHR),} \\
where \textit{LLM Text Summarizer} is DeepSeek-R1 and \textit{LLM Encoder} is Qwen3-Embedding. \\

\noindent\textbf{\textit{Supervised Fine-Tuned LLM (LoRA) Pipeline:}} \\
\texttt{EHR Data → Medical Text → LLM Text Summarization → LLM Encoding → LoRA Adaptation → P(Cancer|EHR),} \\
where \textit{LLM Text Summarizer} is DeepSeek-R1 and \textit{LLM Encoder} is Qwen3-Embedding. \\

\subsection{Can-SAVE Method}

The Can-SAVE method is based on a simple but powerful idea: \textit{combining population survival knowledge with machine learning to predict cancer risk in individual patients}. This solution works exclusively with routine EHR. The methodology is specifically designed to address the limitations of real-world deployment while maintaining high predictive performance.

\textbf{Core Hypothesis:} By leveraging survival analysis methods, we can design domain-specific features that significantly improve the ability of a machine learning model to predict cancer risk, combining the statistical rigor of time-to-event modeling with the predictive power of gradient boost.

The method operates through a multi-stage pipeline. Figure~\ref{fig:overview}C illustrates the overall architecture of the model.

\noindent\textbf{Stage 1: Feature Engineering.}
From raw EHR data, we generate more than 700 candidate features that span multiple domains to capture diverse risk signals:
\begin{itemize}
    \item \textit{Socio-demographic features}: age, sex, BMI, temporal characteristics;
    \item \textit{Temporal patterns}: visit frequency, seasonality effects, time intervals between healthcare encounters;
    \item \textit{Clinical utilization patterns}: total visits, unique diagnoses or services, recent activity metrics, diagnosis-to-service ratios;
    \item \textit{Medical event frequencies}: occurrence counts per ICD-10 group (e.g. 4 diagnoses in the cardiovascular range I00-I99);
    \item \textit{Temporal dynamics}: time since the first/last visit, average intervals between visits, time since the first occurrence of specific conditions;
    \item \textit{Binary indicators}: presence/absence flags for major disease categories;
    \item \textit{Healthcare utilization signatures}: patterns of medical service consumption in different specialties.
\end{itemize}

\noindent\textbf{Stage 2: Survival Features Construction.}
We construct population-level and personalized survival features (\textit{failure} if cancer detected; otherwise \textit{random right-censoring}) through complementary approaches:
\begin{itemize}
    \item \textit{Population-level patterns (Kaplan-Meier estimators)}:
    \begin{itemize}
        \item Overall population survival probability: $\hat{S}_{KM}^{ALL}(age)$;
        \item Sex-stratified survival probabilities: $\hat{S}_{KM}^{\text{SEX}}(age)$ where \\ $\text{SEX}~\in~\{M, F\}$;
        \item Risk gradient features: $|\hat{S}_{KM}(age) - \hat{S}_{KM}(age+1)|$ where $\hat{S}_{KM}~\in~\{ \hat{S}_{KM}^{ALL}, \hat{S}_{KM}^{\text{SEX}} \}$.
    \end{itemize}
    
    \item \textit{Personalized risk assessment (Accelerated Failure Time model)}: The AFT model captures individual-specific risk trajectories through semiparametric survival regression $S_{AFT}(t) = S_0 \left( \int_{0}^{t} r(x(s);\beta)ds \right)$ where $r(x;\beta)$ is a non-negative function, $S_0(\cdot)$ is a base distribution family, $x$ is a vector consisting of covariates, and $\beta_i$ are estimated parameters by minimizing the likelihood function. This generates:
    \begin{itemize}
        \item Personalized survival probability: $\hat{S}_{AFT}(age)$;
        \item Individual risk gradient: $|\hat{S}_{AFT}(age) - \hat{S}_{AFT}(age+1)|$.
    \end{itemize}
    
\end{itemize}

\noindent\textbf{Stage 3: Gradient Boosting Machine Integration.}
\begin{itemize}
    \item Use the outputs of the survival models in \textbf{Stage~2} as additional features for \textbf{Stage~1};
    \item Train Gradient Boosting Machine (GBM) on the extended set of features for the final prediction.
\end{itemize}

As a result, the Can-SAVE method predicts:
\[P(\text{Cancer}|\text{EHR})=GBM(\text{ML Features} \oplus  \text{Survival Outputs}),\]
where $\oplus$ is a concatenation.

This method allows for the use of both population survival patterns and individual patient characteristics for the most accurate prediction. Survival outputs provide regularization through a priori population knowledge, which improves the generalization of the model according to the bias-variance trade-off theory. In our work, we apply the CatBoost framework as a GBM, since it is resistant to overfitting (critical in medical problems) and interpretable through feature importance.

\subsection{Evaluation}
\textbf{\textit{Primary Metric.}} Problem formulation involves a comparison of risk between two patients. As a result, we solve this problem as a ranking task, aiming to maximize the concentration of high-risk patients at the top of the patient list. To achieve this, we employ Average Precision~\cite{Average_Precision} (AP) as the primary metric that aligns the following considerations: (1)~The AP metric focuses on maximizing the proportion of true positive patients among the total number of selected top patients, thereby maximizing the Precision@TOP; (2)~The AP demonstrates stability even in the presence of extreme class imbalance, as evidenced by its relationship with the AUC PR-curve. For example, in 2023, approximately 250 new standardized cancer cases per 100,000 individuals were diagnosed in Russia~\cite{Russian_Stats}. In addition, we report the ROC AUC score for a possible comparison with other methods.

\noindent\textbf{\textit{Validation Strategy:}} (a) \textit{Training Validation Study:} out-of-sample; (b) \textit{Pilot Validation Study:} out-of-sample \& out-of-time.

% ----------------

\section{Experiments}

In order to quantitatively assess the capabilities of Can-SAVE, we first benchmark its performance against a wide range of alternative methods. We then investigate the factors that enable our approach to achieve these results. Finally, under the supervision of clinical oncologists, we compare the effectiveness of Can-SAVE with the current screening workflow. To address our objectives more precisely, we aim to answer the following research questions:

\begin{itemize}
    \item \textbf{Q1:} Can the Can-SAVE method outperform existing approaches in ranking patients according to cancer risk?
    \item \textbf{Q2:} Do the survival-based variables provide a significant boost to the predictive power of Can-SAVE?
    \item \textbf{Q3:} Which features make the largest contribution to the predictive performance of Can-SAVE?
    \item \textbf{Q4:} How does Can-SAVE behave in a retrospective experiment that closely mirrors real-world conditions, relative to the traditional screening process?
\end{itemize}

\subsection{Numeric Experiments}

\textbf{Dataset.}
To train and validate the models, we have a dataset containing 175,441 patients (18+) for the period 2017-2021. The dataset exclusively contains routine polyclinic (outpatient) data, comprising ICD-10 diagnosis codes and medical service codes, universally available data elements that are available in almost any medical organization. 

We divide the set of patients into several samples in order to perform the correct numerical experiments. To achieve this, we apply the stratification of patients by sex and age. Then, we employ statistical testing to validate the integrity of the data partitioning:
(1)~\textit{Multivariate Two-Sample Test:}~\cite{k_sample_test} $H_0: F_1(x) = ... = F_k(x)$ among all $k$-samples where $F_i(x)$ is a distribution of the age, sex or event frequencies in each of the $k$ groups;
(2)~\textit{Univariate Two-Sample Test:}~\cite{min3_test} $H_0: S_1(t) = S_2(t)$ for each pair of the samples, where $S(t)$ is a time-to-event (cancer detection) distribution across splits;
(3)~\textit{Minimum $p$-value} $> 0.05$ required for all comparisons.

These steps ensure that there are no systematic differences between samples, maintain representativeness in the resulting samples, ensure conclusions, and increase the matching of the Newcastle-Ottawa scale~\cite{NOS_scale} proposed for assessing the high quality of non-randomized studies. The brief characteristics of the resulting samples are represented in Table~\ref{tab:train_samples}.

\begin{table}[h]
\centering
\caption{Main characteristics of the resulting samples}
\begin{tabular}{lcccc}
\multicolumn{1}{c|}{\textbf{Sample}} & \textbf{\begin{tabular}[c]{@{}c@{}}Patient \\ Count\end{tabular}} & \textbf{\begin{tabular}[c]{@{}c@{}}Avg. \\ Age\end{tabular}} & \textbf{\begin{tabular}[c]{@{}c@{}}Male, \\ \%\end{tabular}} & \textbf{\begin{tabular}[c]{@{}c@{}}Cancers \\ (C00-C97)\end{tabular}} \\ \hline
\multicolumn{1}{l|}{Survival Train}  & 12,280                                                            & 41.00                                                        & 40.62                                                        & 212 / 1.73\%                                                          \\
\multicolumn{1}{l|}{Survival Test}   & 12,280                                                            & 41.00                                                        & 39.84                                                        & 196 / 1.60\%                                                          \\
\multicolumn{1}{l|}{Train}           & 70,176                                                            & 40.96                                                        & 40.64                                                        & 1137 / 1.62\%                                                         \\
\multicolumn{1}{l|}{Validate}        & 40,350                                                            & 40.92                                                        & 40.72                                                        & 630 / 1.56\%                                                          \\
\multicolumn{1}{l|}{Test}            & 40,355                                                            & 40.97                                                        & 40.51                                                        & 686 / 1.70\%                                                          \\ \hline
\multicolumn{1}{r}{\textbf{Total}}   & \textbf{175,441}                                                  & \textbf{40.96}                                               & \textbf{40.57}                                               & \textbf{2 861 / 1.63\%}                                              
\end{tabular}
\label{tab:train_samples}
\end{table}

\textbf{Survival Models Training.} 
\textit{Kaplan-Meier estimators:} Using the Survival Train and Survival Test samples, the following Kaplan-Meier estimators were fitted: $\hat{S}_{KM}^{ALL}(t)$ for both males \& females, $\hat{S}_{KM}^{M}(t)$ for males, and $\hat{S}_{KM}^{F}(t)$ for females. Detailed information on the fitted Kaplan-Meier estimators can be found in Appendix~\ref{apendix_KM}.

\textit{AFT model:} We trained the AFT model on the Survival Train sample, using the lifelines framework with 100 Optuna optimization trials, and then validated the AFT model on the Survival Test sample. Detailed information on the fitted AFT estimators can be found in Appendix~\ref{apendix_AFT}.

\textbf{Comparison with Baselines.} We performed a numerical experiment to compare the Can-SAVE method versus Baselines. All models were trained on the \textit{Train} sample with hyperparameters optimization performed on the \textit{Validate} sample. Performance of the models tested on the \textit{Test} sample. The results of the experiment are presented in Table~\ref{tab:num_exp}, together with the 95\% confidence intervals.

Across the 17 baselines, Can-SAVE achieved the highest Average Precision (0.228 ± 0.027) surpassing the best fine-tuned LLM and classical/survival methods. Although LoRA-tuned LLM posted the top ROC-AUC (0.901 ± 0.002), its AP remained at 0.193 ± 0.004, underscoring that ROC-AUC alone is insufficient for highly-imbalanced cancer screening. The Can-SAVE method therefore delivers the most effective ranking of high-risk patients without the high computational costs of complex architecture models and LLMs. This experiment allows us to answer \textbf{Q1} positively.

\begin{table}[]
\centering
\caption{Numeric experiment results (\textit{Test} sample; 95\% CI)}
\begin{tabular}{lcc}
\multicolumn{1}{c|}{\textbf{Method}}                                                                                                 & \multicolumn{1}{c|}{\textbf{\begin{tabular}[c]{@{}c@{}}Average \\ Precision\end{tabular}}} & \textbf{ROC AUC}       \\ \hline
\multicolumn{3}{c}{\textit{Machine Learning Pipeline}}                                                                                                                                                                                                          \\ \hline
\multicolumn{1}{l|}{Logistic Regression}                                                                                             & \multicolumn{1}{c|}{0.104 ± 0.013}                                                         & 0.834 ± 0.007          \\
\multicolumn{1}{l|}{Randon Forest}                                                                                                   & \multicolumn{1}{c|}{0.102 ± 0.005}                                                         & 0.833 ± 0.006          \\
\multicolumn{1}{l|}{GBM (CatBoost)}                                                                                                  & \multicolumn{1}{c|}{0.160 ± 0.018}                                                         & 0.786 ± 0.013          \\ \hline
\multicolumn{3}{c}{\textit{Survival Models Pipeline}}                                                                                                                                                                                                               \\ \hline
\multicolumn{1}{l|}{AFT model}                                                                                                       & \multicolumn{1}{c|}{0.117 ± 0.017}                                                         & 0.848 ± 0.022          \\
\multicolumn{1}{l|}{Randon Survival Forest}                                                                                          & \multicolumn{1}{c|}{0.074 ± 0.003}                                                         & 0.786 ± 0.005          \\
\multicolumn{1}{l|}{DeepHit}                                                                                          & \multicolumn{1}{c|}{0.102 ± 0.025}                                                         & 0.864 ± 0.016          \\
\multicolumn{1}{l|}{Deep Survival Machines}                                                                                          & \multicolumn{1}{c|}{0.101 ± 0.005}                                                         & 0.823 ± 0.006          \\ \hline
\multicolumn{3}{c}{\textit{Deep Learning (RNN \& Transformers) Pipeline}}                                                                                                                                                                                                           \\ \hline
\multicolumn{1}{l|}{Fine-tuned CoLES}                                                                                                & \multicolumn{1}{c|}{0.103 ± 0.002}                                                         & 0.813 ± 0.002          \\
\multicolumn{1}{l|}{BERT -\textgreater GRU}                                                                                          & \multicolumn{1}{c|}{0.151 ± 0.026}                                                         & 0.849 ± 0.008          \\
\multicolumn{1}{l|}{Longformer -\textgreater GBM}                                                                                    & \multicolumn{1}{c|}{0.093 ± 0.002}                                                         & 0.777 ± 0.005          \\ \hline
\multicolumn{3}{c}{\textit{LLM Encoding Pipeline}}                                                                                                                                                                                                                  \\ \hline
\multicolumn{1}{l|}{Qwen3-Emb -\textgreater GBM}                                                                                     & \multicolumn{1}{c|}{0.151 ± 0.009}                                                         & 0.869 ± 0.003          \\
\multicolumn{1}{l|}{Qwen3-Emb -\textgreater DeepHit}                                                                                     & \multicolumn{1}{c|}{0.186 ± 0.007}                                                         & 0.885 ± 0.003          \\
\multicolumn{1}{l|}{DeepSeek-R1 -\textgreater GBM}                                                                                   & \multicolumn{1}{c|}{0.164 ± 0.010}                                                         & 0.873 ± 0.005          \\
\multicolumn{1}{l|}{GigaChat -\textgreater GBM}                                                                                      & \multicolumn{1}{c|}{0.185 ± 0.002}                                                         & 0.896 ± 0.001          \\ \hline
\multicolumn{3}{c}{\textit{LLM Summarization \& Encoding Pipeline}}                                                                                                                                                                                                 \\ \hline
\multicolumn{1}{l|}{\begin{tabular}[c]{@{}l@{}}DeepSeek-R1 -\textgreater \\ -\textgreater Qwen3-Emb -\textgreater GBM\end{tabular}}  & \multicolumn{1}{c|}{0.176 ± 0.010}                                                         & 0.881 ± 0.005          \\ %\hline
\multicolumn{1}{l|}{\begin{tabular}[c]{@{}l@{}}DeepSeek-R1 -\textgreater \\ -\textgreater Qwen3-Emb -\textgreater DeepHit\end{tabular}}  & \multicolumn{1}{c|}{0.174 ± 0.004}                                                         & 0.895 ± 0.002          \\ \hline
\multicolumn{3}{c}{\textit{Supervised Fine-Tuned LLM (LoRA) Pipeline}}                                                                                                                                                                                              \\ \hline
\multicolumn{1}{l|}{\begin{tabular}[c]{@{}l@{}}DeepSeek-R1 -\textgreater \\ -\textgreater Qwen3-Emb -\textgreater LoRA\end{tabular}} & \multicolumn{1}{c|}{0.193 ± 0.004}                                                         & \textbf{0.901 ± 0.002} \\ \hline
\multicolumn{3}{c}{\textit{Proposed Method}}                                                                                                                                                                                                               \\ \hline
\multicolumn{1}{l|}{Can-SAVE}                                                                                                        & \multicolumn{1}{c|}{\textbf{0.228 ± 0.027}}                                                & 0.837 ± 0.017          \\ \hline
\end{tabular}
\label{tab:num_exp}
\end{table}

\subsection{Ablation Study}
The ablation analysis shows that isolating the two components of Can-SAVE, the standalone GBM and the survival model AFT, confirms their complementary functions: GBM alone achieves an Average Precision of $0.160 \pm 0.018$, while AFT alone reaches just $0.117 \pm 0.017$. When survival outputs are fused with GBM in Can-SAVE, the Average Precision rises to $0.228 \pm 0.027$ without sacrificing ROC-AUC, demonstrating that survival-derived signals supply critical ranking power that GBM or AFT could not deliver in isolation. These results answer \textbf{Q2}.

\subsection{Feature Importance}
We study the features incorporated in the final model of Can-SAVE. To achieve this, we compute 
\begin{itemize}
    \item \textit{CatBoost Feature Importance} (denoted as \textbf{FI});
    \item \textit{Permutation Importance} (denoted as \textbf{PI}) for Average Precision with 5 times of Monte Carlo replications).
\end{itemize}
From 700 features, we selected factors with CatBoost Feature Importance is $\ge 1$. The remaining features have a much weaker effect on model predictions and were removed.
The predictive power of Can-SAVE is based on several significant factors. \textit{Age} dominates (\textbf{20.2}), which is consistent with cancer epidemiology.
The \textit{output of survival models} contributes \textbf{39.6} of total importance: population-level curves (12.9), sex-specific patterns (9.9) that capture risk trajectories, and the risk gradient that quantifies risk acceleration (14.4).
\textit{Visit pattern} features contribute \textbf{21.5}.
\textit{Clinical markers} (\textbf{17.4}) include immune system services (6.4), suggesting immune dysfunction, and diagnosis of benign neoplasms (6.8), which may reflect the development of precancerous disease.
This interpretable feature hierarchy supports our integration for clinical deployment and addresses \textbf{Q3}.

Although the study relies on a domestic Russian service-coding scheme, Tables~\ref{tab:feature_importance} and~\ref{tab:aft_model} indicate that service-related variables contribute only marginally to the predictive performance of Can-SAVE; moreover, both features employed ("Frequency of medical services for the Immune system" and "Service visits / All visits") can be reproduced in any alternative medical-service coding system.

\begin{table}[h]
\centering
\caption{Feature Importance for the Can-SAVE method (the features are divided into four groups: sociodemographic parameters, survival models, patterns of visits, and clinical markers)}
\begin{tabular}{clcc}
\multicolumn{1}{c|}{\textbf{\#}} & \multicolumn{1}{c|}{\textbf{Feature}}                                                                           & \textbf{FI} & \textbf{PI} \\ \hline
\multicolumn{1}{c|}{1}          & \multicolumn{1}{l|}{Age of the patient}                                                                           & 20.218       & 2.168         \\
\multicolumn{1}{c|}{2}         & \multicolumn{1}{l|}{Sex of the patient}                                                                         & 1.573        & 0.122         \\ \hline

\multicolumn{1}{c|}{3}          & \multicolumn{1}{l|}{$\hat{S}^{ALL}_{KM}(age)$}                                                                  & 12.993       & 1.917         \\
\multicolumn{1}{c|}{4}          & \multicolumn{1}{l|}{$\hat{S}^{SEX}_{KM}(age)$}                                                                  & 9.927        & 2.337         \\
\multicolumn{1}{c|}{5}          & \multicolumn{1}{l|}{$|\hat{S}^{ALL}_{KM}(age+1) - \hat{S}^{ALL}_{KM}(age)|$}                                    & 6.995        & 1.790         \\
\multicolumn{1}{c|}{6}          & \multicolumn{1}{l|}{$|\hat{S}^{SEX}_{KM}(age+1) - \hat{S}^{SEX}_{KM}(age)|$}                                    & 3.842        & 0.337         \\
\multicolumn{1}{c|}{7}         & \multicolumn{1}{l|}{$|\hat{S}_{AFT}(age+1)-\hat{S}_{AFT}(age)|$}                                                & 3.648        & 0.123         \\
\multicolumn{1}{c|}{8}         & \multicolumn{1}{l|}{$\hat{S}_{AFT}(age)$}                                                                       & 2.284        & 0.090         \\ \hline

\multicolumn{1}{c|}{9}          & \multicolumn{1}{l|}{Weeks after first visit}                                                                    & 7.265        & 3.847         \\
\multicolumn{1}{c|}{10}          & \multicolumn{1}{l|}{Month of the visit}                                                              & 7.004        & 2.273         \\
\multicolumn{1}{c|}{11}          & \multicolumn{1}{l|}{Diagnose Visits / All Visits}                                                               & 3.801        & 0.170         \\
\multicolumn{1}{c|}{12}         & \multicolumn{1}{l|}{Service Visits / All Visits}                                                                & 3.489        & 0.032         \\ \hline

\multicolumn{1}{c|}{13}          & \multicolumn{1}{l|}{\begin{tabular}[c]{@{}l@{}}Frequency of  medical services\\ for Immune system\end{tabular}} & 6.390        & 0.112         \\
\multicolumn{1}{c|}{14}         & \multicolumn{1}{l|}{\begin{tabular}[c]{@{}l@{}}Time from the first \\ occurrence of D00-D48\end{tabular}}       & 3.679        & 0.097         \\
\multicolumn{1}{c|}{15}         & \multicolumn{1}{l|}{Frequency of D37-D48}                                                                       & 3.159        & 1.248         \\
\multicolumn{1}{c|}{16}         & \multicolumn{1}{l|}{\begin{tabular}[c]{@{}l@{}}Time from the first \\ occurrence of I00-I99\end{tabular}}       & 1.917        & 0.043         \\
\multicolumn{1}{c|}{17}         & \multicolumn{1}{l|}{Frequency of O20-O29}                                                                       & 1.354        & 0.018         \\
\multicolumn{1}{c|}{18}         & \multicolumn{1}{l|}{\begin{tabular}[c]{@{}l@{}}Time from the first \\ occurrence of Q00-Q99\end{tabular}}       & 0.996        & 0.024         \\ 

%\multicolumn{4}{l}{$^1$ CatBoost Feature Importance}                                                                                                                                \\
%\multicolumn{4}{l}{$^2$ Permutation Importance}                                                                                                                         
\end{tabular}
\label{tab:feature_importance}
\end{table}

\subsection{Oncologist-Supervised Retrospective Study}

\textbf{Design of Experiment.}
The goal of the experiment is to validate the Can-SAVE method in conditions as close to real as possible. To achieve this (Figure~\ref{fig:overview}B):
\begin{enumerate}
    \item Assess the risk of each patient of population using Can-SAVE;
    \item Form a risk group from the Top 1,000 patients;
    \item Pass the list of risk group patients to supervised oncologists;
    \item Supervised oncologists verify the number of correct patients (diagnosed with cancer);
    \item Compare with the traditional examination (baseline).
\end{enumerate}

\textbf{Dataset.}
Our evaluation uses data from 1.9 million patients in five Russian regions (Table~\ref{tab:data_retro}), representing one of the largest cancer prediction validation studies, and spans 2016-2023. 
The criteria for the inclusion of patients were as close as possible to the actual state of affairs of the information stored in medical institutions. Patients were included in the study if they met all the following criteria:
(a)~Age $\ge 18$ years at the time of risk assessment;
(b)~History of no cancer diagnosis (ICD-10 codes C00-C97);
(c)~Not participated in Can-SAVE training.
(d)~The patient has to be included even if his EHR is empty, because the state guarantees the right to preventive protection for everyone.

\begin{table}[h]
\centering
\caption{Dataset for oncologist-supervised retrospective study}
\begin{tabular}{c|cccl}
\textbf{Region} & \textbf{Population} & \textbf{Male, \%} & \textbf{Period} & \multicolumn{1}{c}{\textbf{$t_{pred}$}} \\ \hline
A               & 93 000              & 37\%              & 2020-2021       & 2022/01/01                              \\
B               & 112 620             & 43\%              & 2020-2021       & 2022/01/01                              \\
C               & 165 355             & 32\%              & 2021-2022       & 2023/01/01                              \\
D               & 651 697             & 44\%              & 2016-2017       & 2018/01/01                              \\
E               & 889 293             & 44\%              & 2022-2023       & 2024/01/01                              \\ \hline
                & \textbf{1 911 965}  & \textbf{43\%}     &                 &                                        
\end{tabular}
\label{tab:data_retro}
\end{table}

\textbf{Results.}
For \textbf{Q4}, the results in Table~\ref{tab:results_retro} show that in five Russian regions of 1.9 million patients, Can-SAVE captured 41-90 cancers per 1,000 high-risk patients, versus 9-15 per 1,000 under existing screening, providing a boost of 4.1x-10.0x at identical resource levels. This consistent outperformance was sustained despite the wide variation in population size, chronology, and EHR systems, underscoring the robustness and portability of the model. Moreover, this result is consistently maintained for different age groups as shown in Appendix~\ref{apendix_AgeGroup}. By transforming routine ICD-10 and service codes into actionable risk rankings, Can-SAVE demonstrably unlocks population-scale early detection capacity unattainable with traditional protocols alone.

\begin{table}[h]
\centering
\caption{Results of the oncologist-supervised retrospective experiments in five regions of Russia (1,000 patients in each risk group)}
\begin{tabular}{c|ccc}
\textbf{Region} & \textbf{\begin{tabular}[c]{@{}c@{}}Traditional \\ Examination\end{tabular}} & \textbf{Can-SAVE} & \textbf{Uplift} \\ \hline
A               & 10                                                                         & 41                & 4.1x            \\
B               & 10                                                                          & 58                & 5.8x            \\
C               & 15                                                                          & 71                & 4.7x            \\
D               & 13                                                                          & 84                & 6.5x            \\
E               & 9                                                                           & 90                & 10.0x             %\\ \hline
\end{tabular}
\label{tab:results_retro}
\end{table}

% ----------------

\section{Implementation in a Clinical Setting}
We implemented Can-SAVE in the \textbf{12-month prospective pilot supervised by oncologists} that covered the entire adult population of the Russian region (426,210 patients without prior cancer diagnosis ICD-10 C00–C97 at the start). The deployment was integrated into the existing public health infrastructure with the backend architecture described in Section~\ref{prospect_Backend}.

\subsection{Backend Architecture} \label{prospect_Backend}
\begin{figure}[h]
  \includegraphics[width=8.5cm]{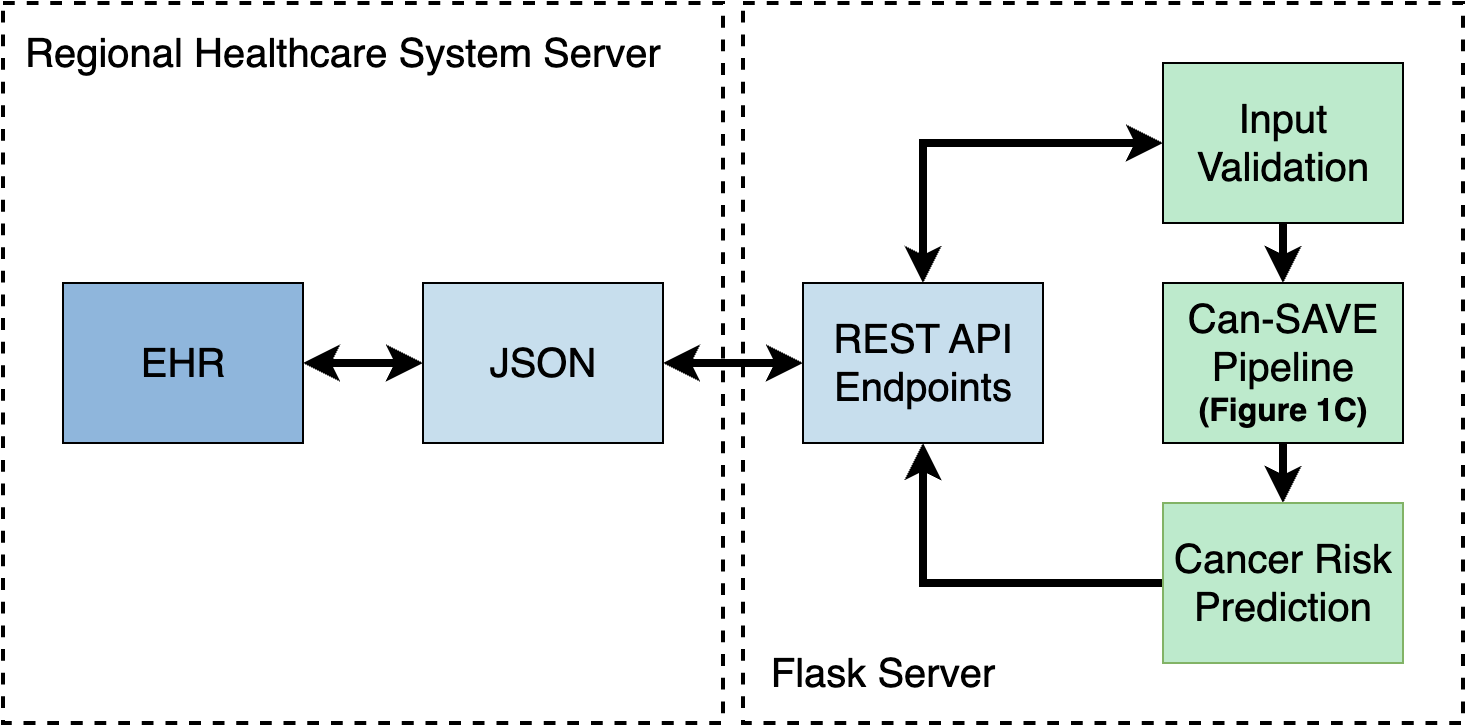}
  \caption{Backend architecture of the Can-SAVE deployment}
  \label{fig:backend}
\end{figure}

Figure~\ref{fig:backend} illustrates the inference of a microservice based on the Flask web application framework:
(1) The REST API receives a JSON-object containing the patient's EHR;
(2) The patient's EHR then calls the Can-SAVE pipeline (Figure~\ref{fig:overview}C) to assess the risk of malignant neoplasms;
(3) As a result, the microservice returns the calculated risk score to clinical users.
This approach enables horizontally scalable deployment in real time. This service operates without storing the processed data, which is critical for processing medical data.

\subsection{Clinical Treatment Process}
The design of the prospective experiment includes the following:
\begin{enumerate}
    \item \textbf{Traditional process (Control):} 320,515 patients scheduled for the national preventive screening (“Dispanserization”) received the standard protocol of examinations mandated by the Russian Ministry of Health;
    \item \textbf{AI-based process (Test):} Can-SAVE ranked a risk-ordered list of 320,515 patients (the same length for fair comparison). Figure~\ref{fig:overview}D illustrates the workflow of this process. Each high-risk patient receives: (a) primary care consultation focused on oncological vigilance (symptom checklist, family history, risk factor questionnaire); (b) immediate referral to an oncologist when indicated.
\end{enumerate}

Both processes were carried out concurrently as a real-world deployment study, allowing a rigorous post-launch evaluation of the clinical impact of the AI system. Invitations, attendances, and confirmed malignancies were recorded automatically.

\textbf{Clinical Results.} Of the 320,515 to be inspected, only 131,167 (40.9\%) attended at least one visit in either arm during the evaluation window. The overlap in attendance between lists was 49.5\%, demonstrating that Can-SAVE surfaces a large subset of patients who would not have been called by the traditional process. The detailed results are presented in Table~\ref{tab:result_prospect}.

\begin{table}[h]
\centering
\caption{Quantification of post-deployment clinical performance: 12-month prospective pilot (426,210 patients)}
\begin{tabular}{lccc}
\multicolumn{1}{c|}{\textbf{Metric}}                                                                                 & \textbf{\begin{tabular}[c]{@{}c@{}}Traditional \\ (Control)\end{tabular}} & \textbf{\begin{tabular}[c]{@{}c@{}}Can-SAVE \\ (Test)\end{tabular}} & \textbf{Uplift} \\ \hline
\multicolumn{1}{l|}{Invited Patients}                                                                                   & 320,515                                                                             & 320,515*                                                            &       --          \\
\multicolumn{1}{l|}{Patients Actually Seen}                                                                             & 131,167                                                                             & 131,167*                                                            &        --         \\ \hline
\multicolumn{1}{l|}{\begin{tabular}[c]{@{}l@{}}Detected Cancers \\ (C00-C97)\end{tabular}}                              & 1,123                                                                               & 2,148                                                               & \textbf{+91\%}           \\
\multicolumn{1}{l|}{\begin{tabular}[c]{@{}l@{}}Detection Rate \\ (per 1,000)\end{tabular}}                    & 8,56                                                                                & 16,38                                                               & \textbf{1.9x}            \\ \hline
\multicolumn{1}{l|}{\begin{tabular}[c]{@{}l@{}}Cancers Coverage \\ (Total Cancers = 2,850)\end{tabular}} & 39,4\%                                                                              & 75,4\%                                                              & \textbf{+36 pp}          \\ \hline
\multicolumn{4}{l}{* \textit{matched 1:1 with the Control cohort to remove coverage bias}}                                                                                                                                                                                                               
\end{tabular}
\label{tab:result_prospect}
\end{table}

It allows us to draw the following conclusions about the clinical impact:
(1)~Can-SAVE nearly \textbf{\textit{doubled}} the cancers found with the same clinical capacity (2,148 vs 1,123), demonstrating superior patient prioritization without additional costs;
(2)~The AI workflow \textbf{\textit{identified 75\% of all cancers}} detected in the region that year, versus 39\% for routine screening;
(3)~\textbf{\textit{No new hardware or type of examinations}} were required. The only change was a risk-prioritized invitation and a brief questionnaire at the therapist;
(4)~Results of the \textbf{\textit{detected nosologies}}, \textbf{\textit{age groups}}, and \textbf{\textit{comparison with specialized screenings}} (Appendix~\ref{apendix_prospect}) also demonstrate the superiority of the AI workflow.

\textbf{Ethical Considerations.}
The conduct of the prospective pilot complies fully with all the requirements of the ethical policy, since all patients in both the Control and Test groups received medical care according to the already approved regulations and protocols of the Russian Federation Ministry of Health (Dispanserization: Order No. 404n, 2021; Oncological alertness (workflow): Order No. 116n, 2021 and Order No. 142n, 2024).

\subsection{System Scalability and Performance}
To make a decision on the deployment of Can-SAVE, the computational cost of such a system is estimated.
We estimate maximum system performance for the population in the range from 10K to 100M, where each EHR contains 100 medical events (close to maximum). Performance evaluation is conducted on the following hardware: Intel® Core™ i7-12700H, 64GB RAM, 1TB ROM. The results are presented in Table~\ref{tab:performance}.

\begin{table}[h]
\centering
\caption{Resource requirements of the Can-SAVE deployment for various size of population}
\begin{tabular}{c|cccc}
\textbf{Patients} & \textbf{Scale} & \textbf{\begin{tabular}[c]{@{}c@{}}Traffic \\ Received\end{tabular}} & \textbf{\begin{tabular}[c]{@{}c@{}}Traffic \\ Sent\end{tabular}} & \textbf{\begin{tabular}[c]{@{}c@{}}Can-SAVE\\ Evaluation\end{tabular}} \\ \hline
10K               & Town              & 2.2MB                                                                & 0.2MB                                                            & 1.8 min                                                                \\
1M                & City              & 221.3MB                                                              & 16.2MB                                                           & 2.9 hours                                                              \\
100M              & Country           & 21.6GB                                                               & 1.6GB                                                            & 12.2 days                                                             
\end{tabular}
\label{tab:performance}
\end{table}

% ----------------

\section{Conclusion}
Can-SAVE demonstrates that pragmatic AI, grounded in survival analysis and trained on ubiquitous EHR codes, can transform population cancer screening without additional hardware, biomarkers, or other specific data in the EHR. Deployed on a national scale it 
(1) increases detection efficiency by up to 10 times in retrospective experiments, 
(2) nearly doubles real-world case-finding while expanding coverage by 36 percentage points, and 
(3) operates within existing primary care workflows on commodity servers. 
Quantification of the post-launch performance (Table~\ref{tab:result_prospect}) confirms that the deployed AI-for-medicine system maintains its efficacy under routine operations.

Future work should extend Can-SAVE to multi-year horizons and incorporate hospital data, but the present study already provides a reproducible template and compelling evidence for health systems seeking cost-effective, scalable AI screening tools.

% ----------------

%\begin{acks}
%To Robert, for the bagels and explaining CMYK and color spaces.
%\end{acks}

\bibliographystyle{ACM-Reference-Format}
%\bibliography{sample-base}

\newpage
\appendix
\section{Supplementary Material}

\subsection{Fitted Kaplan-Meier Estimators} \label{apendix_KM}
\begin{figure}[h]
  \centering
  \includegraphics[width=8cm]{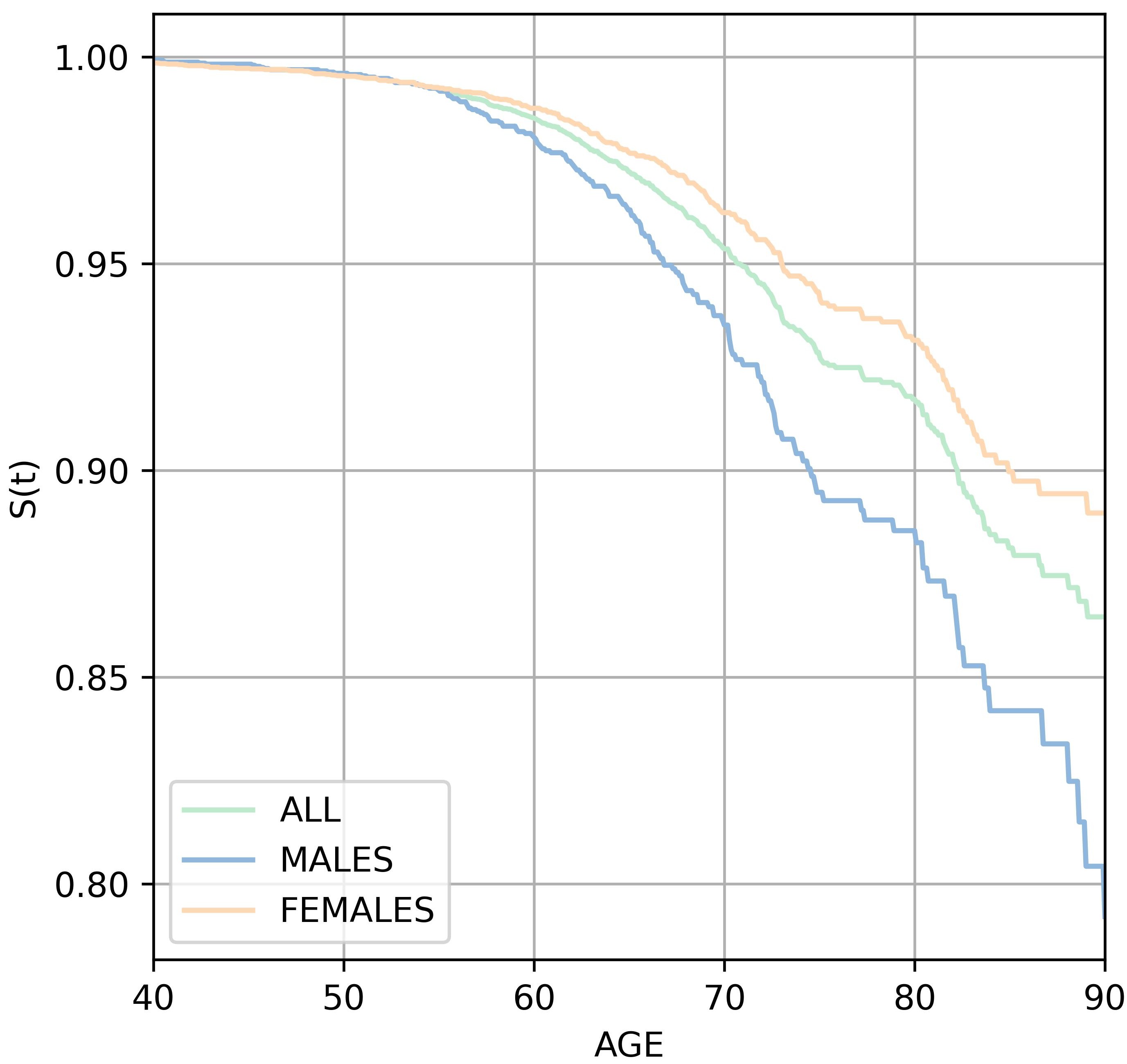}
  \caption{The fitted Kaplan-Meier estimators for males (blue), females (red), and all patients (green)}
  \label{fig:kaplan_meier}
\end{figure}

Figure~\ref{fig:kaplan_meier} demonstrates the resulting Kaplan-Meier estimators. The survival curve for males (blue line) is below the survival curve for females (red line), which is consistent with published statistical data~\cite{Russian_Stats}.

\subsection{Fitted AFT model} \label{apendix_AFT}
The fitted AFT model with $S_0(\cdot)$ based on the Weibull family distribution and $r(x;\beta)=\exp \left( \beta_0+\sum_j \beta_jx_j \right)$ reaches 0.83 of the $C$-Index and 4713.64 of the Akaike Information Criterion (AIC). Furthermore, the log-likelihood ratio test ($-1546.47$) confirms that the fitted model is preferable to the alternative model (without covariates). Table~\ref{tab:aft_model} presents the covariates of the AFT model, the regression coefficients, and the statistics of the $z$-test. All covariates are strictly significant and included in the model with a significance level less than 0.005.

\begin{table}[h]
\centering
\caption{Coefficients of the AFT model (significance $<0.005$)}
\begin{tabular}{l|ccc}
\multicolumn{1}{c|}{\textbf{Covariate (Feature)}} & \textbf{Type} & \textbf{Coef.} & \textbf{$z$-test} \\ \hline
Sex (1 - M, 0 - F)                                & Binary        & 6,64           & 12,68             \\
Binary Indicator (D00-D48)                        & Binary        & -6,71          & -11,34            \\
Binary Indicator (I00-I99)                        & Binary        & 2,62           & 6,26              \\
Binary Indicator (N40-N51)                        & Binary        & -8,08          & -6,88             \\
Service Visits / All Visits                       & Float         & 10,22          & 12,61             \\
Weeks After First Visit                           & Float         & 0,11           & 13,95             \\
Avg. Weeks Between Visits                         & Float         & 1,61           & 12,88             \\
Intercept                                         & Float         & -1,46          & -33,49           \\ %\hline
\end{tabular}
\label{tab:aft_model}
\end{table}

\subsection{Oncologist-Supervised Retrospective Study} \label{apendix_AgeGroup}
\textbf{Age-Groups.}
Table~\ref{tab:AgeGroup} demonstrates the results of the retrospective experiments for each age group between traditional examinations (Control) and Can-SAVE (Test). It can be seen that regardless of the region of Russia and age group, the Can-SAVE method is superior to the age-gender baseline. This allows us to draw a conclusion about the stability and viability of the Can-SAVE method, which is very important when applied to solving problems in the medical domain. This also means that Can-SAVE is able to successfully solve the AI screening task for each age group separately.

\begin{table}[h]
\centering
\caption{Comparison of cancer detection for various age-group during oncologist-supervised retro-experiment}
\begin{tabular}{c|c|ccccc}
\textbf{Region}    & \textbf{Method} & \textbf{35-45} & \textbf{45-55} & \textbf{55-65} & \textbf{65-75} & \textbf{75+}  \\ \hline
\multirow{2}{*}{A} & Traditional     & 0.76           & 1.50           & 3.15           & 4.61           & 4.59          \\
                   & Can-SAVE        & \textbf{1.80}  & \textbf{2.70}  & \textbf{5.90}  & \textbf{8.20}  & \textbf{5.40} \\ \hline
\multirow{2}{*}{B} & Traditional     & 0.28           & 0.61           & 1.38           & 2.28           & 2.39          \\
                   & Can-SAVE        & \textbf{0.50}  & \textbf{1.00}  & \textbf{2.90}  & \textbf{4.10}  & \textbf{4.10} \\ \hline
\multirow{2}{*}{C} & Traditional     & 0.26           & 0.47           & 0.89           & 1.25           & 1.36          \\
                   & Can-SAVE        & \textbf{0.30}  & \textbf{0.90}  & \textbf{2.40}  & \textbf{3.00}  & \textbf{2.70} \\ \hline
\multirow{2}{*}{D} & Traditional     & 0.20           & 0.40           & 0.70           & 1.00           & 1.10          \\
                   & Can-SAVE        & \textbf{1.10}  & \textbf{3.50}  & \textbf{8.00}  & \textbf{9.40}  & \textbf{7.40} \\ \hline
\multirow{2}{*}{E} & Traditional     & 0.27           & 0.54           & 1.14           & 1.77           & 1.90          \\
                   & Can-SAVE        & \textbf{0.30}  & \textbf{0.70}  & \textbf{2.60}  & \textbf{5.60}  & \textbf{2.80} \\ \hline
\end{tabular}
\label{tab:AgeGroup}
\end{table}

\subsection{Oncologist-Supervised Prospective Study} \label{apendix_prospect}

\textbf{Detected Nosologies.}
Table~\ref{tab:prospect_Nosologies} shows the structure of the detected malignant neoplasms for both Traditional examinations (Control) and Can-SAVE (Test). As can be seen, Can-SAVE not only quantitatively surpassed the results of the Control group, but also showed consistently high results within each nosological group. All of this allows us to conclude that the AI method is sensitive to the entire spectrum of malignant neoplasms.

\begin{table}[h]
\centering
\caption{Structure of the detected malignant neoplasms under 12-months oncologist-supervised prospective experiment}
\begin{tabular}{lccc}
\multicolumn{1}{c|}{\textbf{\begin{tabular}[c]{@{}c@{}}Malignant \\ Neoplasms of\end{tabular}}}          & \textbf{ICD-10}             & \textbf{\begin{tabular}[c]{@{}c@{}}Traditional \\ (Control)\end{tabular}} & \textbf{\begin{tabular}[c]{@{}c@{}}Can-SAVE \\ (Test)\end{tabular}} \\ \hline
\multicolumn{1}{l|}{\begin{tabular}[c]{@{}l@{}}Lip, oral cavity \\ and pharynx\end{tabular}}             & C00-C14                     & 37                                                                        & 51                                                                  \\
\multicolumn{1}{l|}{Digestive organs}                                                                    & C15-C26                     & 287                                                                       & 570                                                                 \\
\multicolumn{1}{l|}{\begin{tabular}[c]{@{}l@{}}Respiratory and \\ intrathoracic organs\end{tabular}}     & C30-C39                     & 121                                                                       & 276                                                                 \\
\multicolumn{1}{l|}{\begin{tabular}[c]{@{}l@{}}Bone and articular \\ cartilage\end{tabular}}             & \multicolumn{1}{l}{C40-C41} & 4                                                                         & 2                                                                   \\
\multicolumn{1}{l|}{Skin}                                                                                & C43-C44                     & 140                                                                       & 308                                                                 \\
\multicolumn{1}{l|}{\begin{tabular}[c]{@{}l@{}}Mesothelial \\ and soft tissue\end{tabular}}              & C45-C49                     & 8                                                                         & 14                                                                  \\
\multicolumn{1}{l|}{Breast}                                                                              & C50                         & 148                                                                       & 222                                                                 \\
\multicolumn{1}{l|}{\begin{tabular}[c]{@{}l@{}}Genitourinary \\ system\end{tabular}}                     & C51-C68                     & 274                                                                       & 512                                                                 \\
\multicolumn{1}{l|}{Eye, brain, CNS}                                                                     & C69-C72                     & 10                                                                        & 22                                                                  \\
\multicolumn{1}{l|}{\begin{tabular}[c]{@{}l@{}}Thyroid and \\ endocrine glands\end{tabular}}             & C73-C75                     & 36                                                                        & 52                                                                  \\
\multicolumn{1}{l|}{\begin{tabular}[c]{@{}l@{}}Ill-defined, secondary \\ unspecified sites\end{tabular}} & C76-C80                     & 19                                                                        & 33                                                                  \\
\multicolumn{1}{l|}{\begin{tabular}[c]{@{}l@{}}Lymphoid, \\ haematopoietic, etc.\end{tabular}}           & C81-C96                     & 38                                                                        & 83                                                                  \\
\multicolumn{1}{l|}{\begin{tabular}[c]{@{}l@{}}Independent\\ multiple sites\end{tabular}}                & C97                         & 1                                                                         & 3                                                                   \\ \hline
\textbf{}                                                                                                & \textbf{}                   & \textbf{1,123}                                                            & \textbf{2,148}                                                     
\end{tabular}
\label{tab:prospect_Nosologies}
\end{table}

\textbf{Age Groups.}
Table~\ref{tab:prospect_AgeGroups} shows the detection of malignant neoplasms in different age groups of patients for both Traditional examinations (Control) and Can-SAVE (Test). It can be seen that the age structure of the patients selected by Can-SAVE differs from the Control Group. This can be explained by a number of reasons, including the purpose of the dispanserization to search for not only oncological diseases but also other chronic diseases. However, it should be noted that the concentration of oncological patients in the Test Group is significantly higher. This allows the Can-SAVE method to be used for different scenarios, for example, to form an additional group of patients of a certain age and gender composition.

\begin{table}[h]
\centering
\caption{Detection of malignant neoplasms in different age groups under 12-months oncologist-supervised prospective experiment}
\begin{tabular}{cccc}
\multicolumn{1}{c|}{\textbf{\begin{tabular}[c]{@{}c@{}}Age \\ Group\end{tabular}}} & \textbf{\begin{tabular}[c]{@{}c@{}}Patients \\ Seen\end{tabular}} & \textbf{\begin{tabular}[c]{@{}c@{}}Detected \\ Cancers\end{tabular}} & \textbf{\begin{tabular}[c]{@{}c@{}}Detection \\ Rate\end{tabular}} \\ \hline
\multicolumn{4}{c}{\textbf{Traditional Examinations (Control)}}                                                                                                                                                                                                                                    \\ \hline
\multicolumn{1}{c|}{18-39}                                                         & 20,873                                                            & 27                                                                   & 0.13\%                                                             \\
\multicolumn{1}{c|}{40-49}                                                         & 32,553                                                            & 140                                                                  & 0.43\%                                                             \\
\multicolumn{1}{c|}{50-59}                                                         & 27,466                                                            & 201                                                                  & 0.73\%                                                             \\
\multicolumn{1}{c|}{60-69}                                                         & 28,123                                                            & 370                                                                  & 1.32\%                                                             \\
\multicolumn{1}{c|}{70+}                                                           & 22,152                                                            & 385                                                                  & 1.74\%                                                             \\
\multicolumn{1}{r|}{\textbf{Total}}                                                & \textbf{131,167}                                                  & \textbf{1,123}                                                       & \textbf{0.86\%}                                                    \\ \hline
\multicolumn{4}{c}{\textbf{Can-SAVE (Test)}}                                                                                                                                                                                                                                                       \\ \hline
\multicolumn{1}{c|}{18-39}                                                         & 301                                                               & 6                                                                    & 1.99\%                                                             \\
\multicolumn{1}{c|}{40-49}                                                         & 2,355                                                             & 24                                                                   & 1.02\%                                                             \\
\multicolumn{1}{c|}{50-59}                                                         & 25,394                                                            & 239                                                                  & 0.94\%                                                             \\
\multicolumn{1}{c|}{60-69}                                                         & 56,964                                                            & 929                                                                  & 1.63\%                                                             \\
\multicolumn{1}{c|}{70+}                                                           & 46,153                                                            & 950                                                                  & 2.06\%                                                             \\
\multicolumn{1}{r|}{\textbf{Total}}                                                & \textbf{131,167}                                                  & \textbf{2,148}                                                       & \textbf{1.64\%}          \\
\hline
\end{tabular}
\label{tab:prospect_AgeGroups}
\end{table}

\textbf{Comparison with Specialized Screenings.}
We conducted a comparison between the Can-SAVE AI method and existing medical procedures (screenings) in terms of quality. To assess this, we used the Number Needed to Screen (NNS)~\cite{NNS} as a statistical indicator that characterizes the quality of screenings. Table~\ref{tab:NNS} presents the NNS values for Breast, Lung, and Colorectal cancers, along with the corresponding values obtained from the prospective experiment with 131,167 patients. These results show that the number of confirmed cancers identified by Can-SAVE is comparable to the results of medical screenings.

\begin{table}[h]
\centering
\caption{Comparison of cancer detection rates during screenings (NNS) and detected by the Can-SAVE under oncologist-supervised prospective experiment}
\begin{tabular}{c|cccc}
\textbf{Cancer} & \textbf{Age} & \textbf{NNS} & \textbf{\begin{tabular}[c]{@{}c@{}}Cancers \\ per 1000 \\ screenings\end{tabular}} & \textbf{Can-SAVE} \\ \hline
Breast~\cite{NNS_breast}          & 40-79        & 233-746      & 1-4                                                                                & 1.7               \\
Lung~\cite{NNS_lung}            & 18+          & 255-963      & 1-4                                                                                & 2.1               \\
Colorectal~\cite{NNS_colorectal}      & 18-75        & 108-257      & 4-9                                                                                & 4.3              
\end{tabular}
\label{tab:NNS}
\end{table}

\end{document}